\documentclass[preprint,12pt]{elsarticle}

\usepackage{amssymb}

\usepackage{amsmath,empheq}
\usepackage[english]{babel}
\usepackage{amsfonts}
\usepackage{xcolor}
\usepackage{hyperref}
\usepackage{booktabs}
\usepackage{longtable,multirow,multicol}
\usepackage{makecell}
\usepackage[english,ruled,linesnumbered]{algorithm2e}
\usepackage{lscape}
\usepackage{graphicx,array,placeins}

\journal{Computers \& Operations Research}

\begin{document}

\begin{frontmatter}







\title{Enhanced Iterated local search for the technician routing and scheduling problem}

\author[1,2]{Ala-Eddine Yahiaoui}
\author[1]{Sohaib Afifi}
\author[1]{Hamid Allaoui}
\address[1]{Univ. Artois, UR 3926, Laboratoire de Génie Informatique et d'Automatique de l'Artois (LGI2A), Béthune, F-62400, France, e-mail:	\{aeddine.yahiaoui,sohaib.afifi,hamid.allaoui\}@univ-artois.fr}
\address[2]{Universit\'e Laval, Facult\'e des sciences et de g\'enie, 1065 Avenue de la M\'edecine, QC G1V 0A6, e-mail:	ala-eddine.yahiaoui.1@ulaval.ca}

\begin{abstract}
Most public facilities in the European countries, including France, Germany, and the UK, were built during the reconstruction projects between 1950 and 1980. Owing to the deteriorating state of such vital infrastructure has become relatively expensive in the recent decades. A significant part of the maintenance operation costs is spent on the technical staff. Therefore, the optimal use of the available workforce is essential to optimize the operation costs. This includes planning technical interventions, workload balancing, productivity improvement, etc. In this paper, we focus on the routing of technicians and scheduling of their tasks. We address for this purpose a variant of the workforce scheduling problem called the technician routing and scheduling problem (TRSP). This problem has applications in different fields, such as transportation infrastructure (rail and road networks), telecommunications, and sewage facilities. To solve the TRSP, we propose an enhanced iterated local search (eILS) approach. The enhancement of the ILS firstly includes an intensification procedure that incorporates a set of local search operators and removal-repair heuristics crafted for the TRSP. Next, four different mechanisms are used in the perturbation phase. Finally, an \textit{elite set} of solutions is used to extensively explore the neighborhood of local optima as well as to enhance diversification during search space exploration. To measure the performance of the proposed method, experiments were conducted based on benchmark instances from the literature, and the results obtained were compared with those of an existing method. Our method achieved very good results, since it reached the best overall gap, which is three times lower than that of the literature. Furthermore, eILS improved the best-known solution for $34$ instances among a total of $56$ while maintaining reasonable computational times.

\end{abstract}




\end{frontmatter}


\section{Introduction}\label{sec:intro}

Workforce scheduling is a relevant research topic in transportation and logistics, since it can be applied in many fields \cite{castillo2016}, such as technician routing and scheduling, manpower allocation, security personnel routing and rostering and home care services. Interest in this research area is also driven by the importance of ensuring an efficient and satisfying client service policy after a product delivery, which substantially contributes to the maintain of the market share \cite{khalfay2017}. The workforce scheduling problem focuses on the elaboration of models and solution methods for planning in-field personnel activities, including their mobilization between different locations. Moreover, the problem consists in the elaboration of workload allocation and routing of technician crews, as well as the scheduling of their operations at the level of task locations, which include industrial facilities, patient homes, telecommunication infrastructure, etc. In addition, many objectives and challenges may be considered, such as increasing productivity, reducing transportation costs, increasing the number of fulfilled tasks, reducing outsourcing costs, reducing overtime, balancing technician workloads, etc. Furthermore, to have a reliable and satisfactory organization of the workforce in the field, several requirements and constraints have to be met : in addition to the vehicle routing problem classical constraints (capacity and time windows) and work regulations (breaks and workload). Other aspects could be taken into consideration such as skill types and competency levels required by each task, precedence constraints between several tasks for the same customer, priorities, limited crews of technicians, and sometimes the use of specific tools and spare parts.

In this paper, we address a variant of the technician routing and scheduling problem (TRSP) presented by Pillac et al.\cite{pillac2013}. Given a crew of technicians and a set of tasks to fulfill at their respective locations, the goal is to assign subsets of tasks to individual technicians and construct the routes for each technician in such a way that the total duration of the routes is minimized. Several types of constraints must be respected by each route. First, given the multi-depot structure of the TRSP, where each technician is associated with their home depot where they should start and finish the allocated route. A second type of constraints is the compatibility between technicians and tasks, where each task requires proficiency in a specific type of skill by the assigned technician, whereas a given technician may not have necessary the proficiency in all those skills. Both previous conjugated constraints give rise to the site-dependent constraint. Another type of constraints is resource requirement, where each task requires a certain amount of resources of different types. Two general resource classes are considered: tools and spare parts. While the former is renewable, the latter is non-renewable. Moreover, each technician starts the journey from his home depot with a set of tools and an initial inventory of spare parts. However, in the cases where a technician does not have enough tools or spare parts to continue the journey, the inventory can be replenished by visiting a central depot once at some point in the route, where an infinite stock of tools and spare parts is available. Finally, as the TRSP is an extension of the VRPTW, a time window is associated with each task. In addition, each home depot has opening and closing times.

The main contribution of this paper is the proposal of an enhanced iterated local search (eILS) for the TRSP. It incorporates several procedures that are used during the intensification and perturbation phases. First, the intensification phase combines a set of removal-repair heuristics and local search operators. Second, several perturbation procedures are incorporated into the eILS. They differ in whether they are based on remove-repair operators or local search-based deterioration. In both cases, the criteria used may also differ, whether based on travel costs or duration. Hence, four perturbation mechanisms are introduced. On the other hand, an \textit{Elite set} of solutions is used to enhance the intensification and diversity management of the eILS. The extensive intensification approach is inspired by the proximate-optimality principle, and is achieved by allowing each solution to be the starting point of several ILS phases. Whereas diversity management is achieved by maintaining a relatively diverse population of solutions and discarding duplicates solutions that are too close. At the lower level, we propose a new local search operator called \textit{SwapSequence} operator. It interchanges two sequences of $k$ and $k'$ visits between two different routes. A new removal operator is also proposed. It is derived from the related removal operator introduced in \cite{shaw1998}, where instead of removing a set of individual tasks, it removes sequences of tasks. A key feature of the eILS is the implementation of a time-constant moves evaluation and feasibility tests. This includes time window feasibility checks, renewable and nonrenewable resource availability tests, technician skills compatibility tests and duration evaluation. Moreover, several speed-up techniques have been introduced in the eILS to achieve computational efficiency.

The remainder to this paper is organized as follows. A brief state-of-the-art for the general workforce scheduling problem is proposed in Section \ref{sec:liter}. In Section \ref{sec:model} we describe TRSP by introducing the necessary notation as well as an illustrative example. Our metaheuristic approach proposed for TRSP is presented in Section \ref{sec:sol}. The computational tests and detailed results, are presented and discussed in Section \ref{sec:comput}. Finally, we present some conclusions drawn from this study and discuss relevant perspectives and extensions for TRSP.

\section{Related work}\label{sec:liter}

In this paper, we study a variant of the workforce scheduling problem where a set of technicians need to fulfill tasks at different locations. This general class of problems is called the workforce scheduling and routing problem (WSRP) \cite{castillo2016}. A key characteristic of this class of problems is that moving from one location to another takes a significant amount of time, therefore, minimizing travel time will substantially contribute to cost reduction and improved productivity. Applications of these problems in real life can be found in several sectors, such as home health-care services and infrastructure maintenance operations. 

Most WSRP variants found in the literature are extensions for the VRP with time windows (VRPTW). In this class of problems, each point of interest or customer is associated with a time window that specifies when the vehicle should start the service \cite{desrochers1992}. The main objective function of VRPTW is cost minimization. Other variants minimize first the number of vehicles before optimizing travel costs. Another objective function that is less frequently used in the literature is the minimization of the duration \cite{desrochers1988}. In addition to the basic VRPTw, WSRP problems may also consider additional characteristics and constraints, such as multiple depots, multi-trips, site-dependent considerations, etc. In home health care for example, Li et al. presented in \cite{li2021} a variant of the technician routing and scheduling problem with workload balance and outpatient service performed by doctors. Bredstrom \& Ronnqvist. (2007) \cite{bredstrom2007} introduced a new variant of the VRPTW, with additional synchronization constraints between pairs of caregivers on a selected subset of patients. Another variant of the VRPTW with profits and synchronization constraints has also been studied in \cite{yahiaoui2021}, with an application in the context of fire-fighting. 
Several studies in recent years have focused on the integration of uncertainties into the workforce scheduling problem. Chen et al. (2016) \cite{chen2016} considered uncertain service times, and proposed a branch-and-cut approach to solve the problem. Shi et al. (2019) \cite{shi2019} proposed a robust optimization model with uncertain service and travel times and proposed a metaheuristic and Monte Carlo simulation to solve the problem.

One of the first applications of VRP in the field of workforce scheduling was reported by Weigel and Cao. (1999) \cite{weigel1999}. The authors proposed the use of VRPTW to model technician dispatching and home delivery problems faced by a well-known retailer. To solve this problem, they proposed a tabu search algorithm that combines intra and inter-route improvement procedures after the initial assignment of requests to technicians. Xu and Chiu (2001) \cite{Xu2001} addressed the field technician scheduling problem inspired by the telecommunication industry. In the studied problem, the objective is to maximize the preferences when assigning tasks to technicians, as well as the minimization of the work duration. The authors proposed several approaches to solve the problem, namely, a greedy randomized adaptive search procedure (GRASP), upper bounds, relaxation schemes and an extended mathematical model. Tang et al. (2007) \cite{tang2007} modeled a maintenance-scheduling problem on a horizon of several days as several multiple-tour maximum collection problems with time-dependent rewards. The rewards decrease overtime to favor faster scheduling tasks. The authors proposed solving the problem using a tabu search heuristic. Bostel et al. (2008) \cite{bostel2008} addressed a field force planning and routing problem solved on a multi-period time horizon and uses a rolling horizon approach, with an application in the field of water treatment and distribution. A memetic algorithm and a column generation-based heuristic is proposed to deal with the static version. An adapted procedure is proposed to address the dynamic version of the problem. An exact method based on the column generation approach was proposed for the same problem in Tricoire et al. 2013 \cite{tricoire2013}.

The TRSP gained more attention after the French Operations Research Society (ROADEF) dedicated the yearly challenge to addressing the technician and intervention scheduling problem proposed by a well-known telecommunication company \cite{dutot2006}. In this problem, each task is associated with a priority level and requires a certain level of proficiency in a set of skills such that it can be performed by a technician. Technicians are grouped into teams and dispatched to fulfill tasks without the consideration of travel times. The objective of the problem is to execute tasks as early as possible depending on their priority levels. Cordeau et al. (2010) \cite{cordeau2010} developed an ALNS method whereas Hashimoto et al. (2011) \cite{hashimoto2011} proposed a GRASP method. Although the problem does not consider any routing decisions, it has been the origin of several variants of WSRP. Kovacs et al. (2012) \cite{kovacs2012} proposed the service technician routing and scheduling problem (STRSP) which is a generalization of the variant proposed in \cite{dutot2006}, because it takes into consideration the optimization of travel times. Moreover, in the case where it is not possible to execute all the tasks, an outsourcing cost is associated with each unfulfilled task. Two variants are investigated: the first is with team building and the other is without team building. The authors proposed a generic ALNS approach for both variants, which were validated on new benchmark instances derived from the one proposed in \cite{dutot2006}. Later, Xie et al. (2017) \cite{xie2017} tackled the variant without team building in \cite{kovacs2012} and proposed an iterated local search algorithm that succeeded in finding several new best-known solutions for the problem. Mathlouthi et al. (2018) \cite{mathlouthi2018a} introduced a new variant of the TRSP. The main features are the consideration of multiple time windows per task, possibility for technicians to break during the day and possibility of picking up some special types of spare parts that are available only at a subset of central depots. The authors proposed a mathematical formulation that was tested on small-size instances. To tackle large-size instances, the same authors proposed in Mathlouthi et al. (2018) \cite{mathlouthi2018b} a tabu search heuristic enhanced by an adaptive memory mechanism.

Zamorano and Stolletz (2017) \cite{zamorano2017} proposed a new variant of the TRSP that combines team building and multi-period features. This variant was inspired by real-life problems faced by an external maintenance provider specializing in forklifts. The authors proposed a mixed-integer program and branch-and-price algorithm while considering different sub-problem formulations during the column generation phase. Tests are carried out on a set of artificial instances and real-world data. Pekel. (2020) \cite{pekel2020} addressed the same problem presented in \cite{zamorano2017}. The author proposed an improved particle swarm optimization (IPSO) algorithm and compared its results with those of a branch-and-cut algorithm. Guastaroba et al. (2020) \cite{guastaroba2020} presented another variant of the WSRP with multiple periods and team building. To solve this problem, they proposed a mixed-integer program and two meta-heuristic methods. The first method is a math-heuristic approach based on ALNS whereas the second is a three-phase decomposition algorithm.

\section{Problem description}\label{sec:model}

We consider a set of technicians/vehicles $\mathcal{K}=\{0,\ldots,K-1\}$, where $|\mathcal{K}|=K$, and a set of tasks $\mathcal{R}=\{0,\ldots,N-1\}$, where $|\mathcal{R}|=N$. In the following, the two words, technician and vehicle are used interchangeably. Each technician $k \in \mathcal{K}$ starts from its home depot $o_k \in O = \{1+k| k \in K\}$ and ends at the same depot. A central depot "$0$" is open to technicians to replenish their inventory of spare parts and necessary tools. We define $\delta_0$ as the replenishment time at the central depot. Each task $i$ is associated with a location $u_i \in U = \{K+1+i| i \in \mathcal{R}\}$, a service time duration $\delta_i$ and a time window $[e_i,l_i]$, defining earliest and latest service starting times. In addition, for each arc $(i,j)| i,j \in V =  O \cup U \cup \{0\}$, we associate travel time $t_{ij}$, which is the same for all vehicles. We also consider several types of spare parts $T=\{1,\ldots,|T|\}$, tools $P=\{1,\dots,|P|\}$ and skills $Q=\{1,\dots,|Q|\}$. Each task $i \in \mathcal{R}$ requires $d_{ip}$ units of spare parts of type $p \in P$, and the tool $t \in T$ if the Boolean $b_{it}$ is true. We also set the constant $a_{iq}$ to 1 if task $i$ requires skill $q \in Q$. For each vehicle $k \in \mathcal{K}$, we denote the initial inventory of spare parts of type $p \in P$ by $v^k_p$, whereas a Boolean $w^k_t$ is set to 1 if a tool $t \in T$ is in the vehicle starting from the depot. We also set a constant $y^k_{q}$ to 1 if the technician possesses a skill $q \in Q$. Because skills are intrinsic to each technician, we can define a compatibility list of technicians for each task. Hence, we denote for each task $i \in \mathcal{R}$, the set of compatible technicians by $K_i$.
Finally, we associate for each home depot $o_k | k \in \mathcal{K}$ an opening and a closing time window $[e_k,l_k]$.

\section{Enhanced iterated local search}\label{sec:sol}
We provide in this section a detailed description of the eILS. 

\subsection{Low-level heuristics}\label{sec:lowLevelOperators}
In this section, we propose a detailed description of the low-level heuristics/operators used to describe the different components of the eILS. 

\subsubsection{Remove-Repair based perturbation}\label{sec:removeRepaiOperator}
Given an input solution and a perturbation parameter $D_{max}$, this procedure removes a random number of tasks between $1$ and $D_{max}$, and then applies the best insertion algorithm to repair the solution.

\paragraph{\textbf{Best insertion algorithm}}
This heuristic iteratively inserts tasks in the solution at their best positions according to one of the two evaluation criteria : the duration or the travel costs. At each iteration, the unscheduled list of tasks is computed, and all feasible insertions in compatible vehicles are computed for each task. If no feasible insertion for a given task is found in a compatible vehicle, then the best insertion algorithm looks for feasible insertions while considering a pass by the central depot for replenishment. The task with the best insertion is selected and scheduled at its respective position. This process is iterated until all tasks are scheduled or no feasible insertions are found.

\begin{figure}
\begin{center}
\includegraphics[scale=.35]{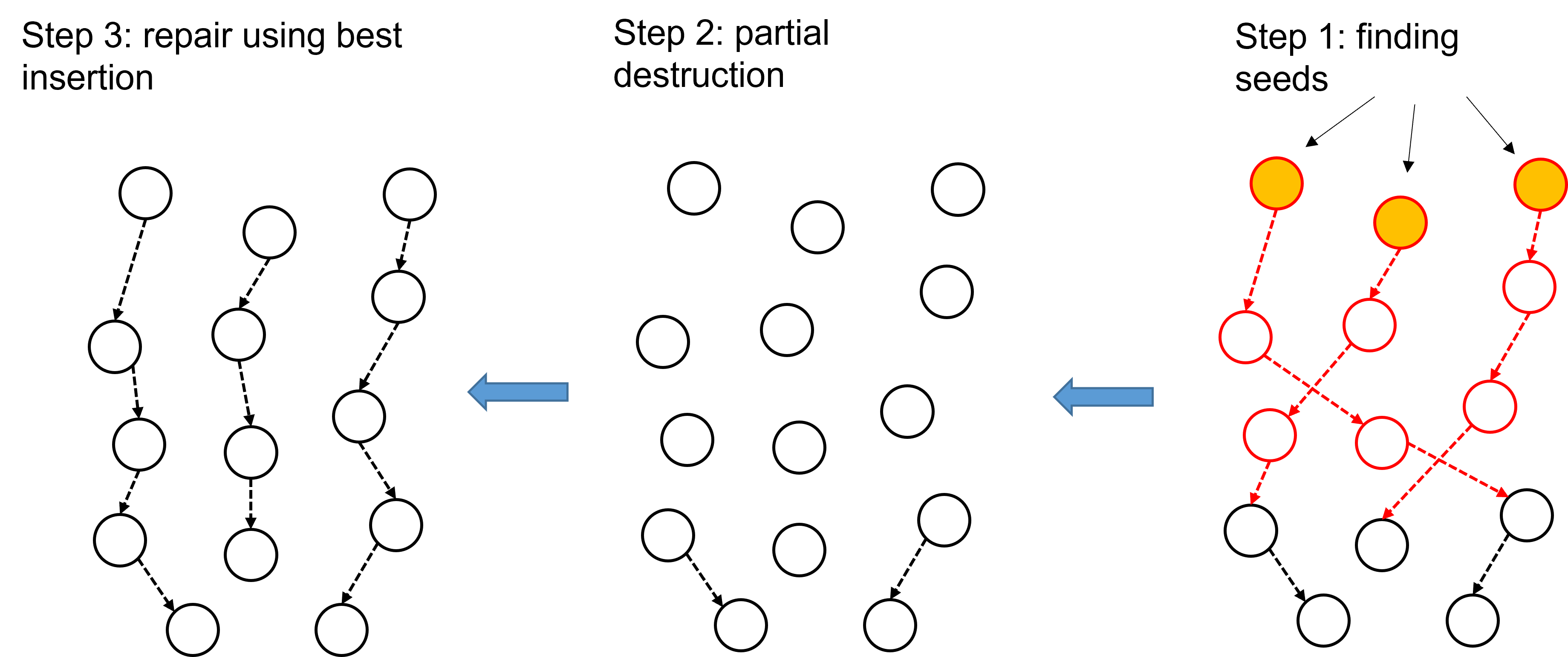}
\end{center}
\caption{Sequence-related removal process}
\label{fig:seqrelat}
\end{figure}

\paragraph{\textbf{Removal operators}}
Given a solution, removal operators remove a random number of tasks between $1$ and $D_{max}$. 
\begin{itemize}
\item \textit{Random Removal} : randomly selects the tasks and removes them from their respective routes.
\item \textit{Worst Removal} : iteratively selects the task yielding the maximum cost reduction. Two versions are available depending on the move evaluation : based on duration or on travel costs.
\item \textit{Sequence related removal} : This removal heuristic is inspired by the \textit{related} removal operator introduced in Shaw et al. \cite{shaw1998} as well as the SISR operator in \cite{christiaens2020}. The basic idea of this operator is to remove sub-sequences of visits from distinct routes so that the best insertion algorithm re-inserts them, and hopefully, constructs promising sub-sequences that improve the objective value of the solution. This approach proved to be efficient, especially on problems where tasks have tight time windows.

In the first step, a set of tasks are selected from different routes using the \textit{related} removal operator of Shaw et al. \cite{shaw1998}. This operator takes into consideration the spatial and temporal relatedness of those tasks. We call these tasks \textit{seeds}. In the second step, a sub-sequence of visits occurring after the \textit{seeds} are removed together with the \textit{seeds}. Fig. \ref{fig:seqrelat} simulates the impact of the \textit{sequence-related} removal operator. The number of removed tasks is initialized by 3, and varies according to the perturbation parameter (See Section \ref{sec:fast}).
\end{itemize}

It is noteworthy to mention that, after every task is removed from a given route, a test is systematically performed to verify whether the visit to the central depot is still relevant, or the initial inventory of tools and spare parts satisfies the requirements of the remaining tasks in the route.

Based on these operators, Algorithm \ref{algo:removalRepairpertur} presents the skeleton of a remove-repair based perturbation procedure.

\begin{algorithm}[!ht]
\SetKwInOut{Input}{input}
\SetKwInOut{Output}{output}
\DontPrintSemicolon
\Input{Solution $S$, perturbation parameter $D_{max}$}
$unscheduled \leftarrow getUnscheduledTasks(S)$\;
$r \leftarrow \mathcal{U}(1,3)$\;
\lIf{$(r = 1)$}{$tmpUnscheduled \leftarrow RandomRemoval(S,D_{max})$}
\lElseIf{$(r = 2)$}{$tmpUnscheduled \leftarrow WorstRemoval(S,D_{max})$}
\lElse{$tmpUnscheduled \leftarrow SeqRelatedRemoval(S,D_{max})$}
$(S,unscheduled) \leftarrow BestInsertion(S,unscheduled)$\;
$(S,tmpUnscheduled) \leftarrow BestInsertion(S,tmpUnscheduled)$\;
$unscheduled \leftarrow unscheduled \cup tmpUnscheduled$\;
\Return $S$
\caption{{\sc Removal/repair perturbation procedure}}\label{algo:removalRepairpertur}
\end{algorithm}

\subsubsection{Local search operators}\label{sec:lsOperators}

The local search procedure is implemented as a variable neighbourhood descent search (VNDS) procedure. This component is composed of two sets of local search operators, inter-route search operators and intra-route search operators.

The inter-route search operators set $\mathcal{N}^e$ comprises three (03) operators.

\begin{itemize}
\item \textit{2-Opt*} : this operator explores the possibilities of exchanging two arcs $(i,j)$ and $(k,l)$ located in two distinct routes by arcs $(i,l)$ and $(k,j)$. Because in the TRSP each vehicle should start and end at the same depot, exchanging arcs between the last visits and the depots should be taken into consideration during moves evaluation.

\item \textit{Swap-relocate} : this operator explores the relocation of a task or an arc to another route, either by placing it between two consecutive visits or by interchanging it with a task in the second route. Moreover, we consider the possibility of reversing the arc before being relocated. This gives raise to six (06) different movements. $Swap-relocate(1,0)$ is a simple relocation of a task from one route to another. $Swap-relocate(1,1)$ is a swap of two tasks visited in two distinct routes. $Swap-relocate(2,0)$ is a relocation of an arc whereas $Swap-relocate(2,1)$ is an interchange of an arc with a task from another route. Finally, $Swap-relocate(2,0)^r$ and $Swap-relocate(2,1)^r$ perform a reverse of the arc before performing the relocation or the interchange. 

\item \textit{Swap-sequence} : this operator is similar to \textit{Swap-relocate}, except that it does not consider the reversal of sub-sequences. Several combinations are possible : \textit{Swap-sequence(2,2)}, \textit{Swap-sequence(3,k)} and \textit{Swap-sequence(4,k)}, where the length of second subsequence $k$ is less than or equal to the first subsequence.

\end{itemize}

Regarding the intra-route search operators set $\mathcal{N}^a$, four (04) operators are considered : 

\begin{itemize}
\item \textit{Exchange$^1$} : this operator explores the possibilities of exchanging the position of two tasks $i$ and $j$ located in the same route.
\item \textit{Shift$^1$} : this operator tries to move a given task $i$ forward and backward to another position in the same route.
\item \textit{R-Opt} : this operator tries to move a sequence of two or three visits forward and backward in the same route.
\item \textit{2-Opt} : this operator explores the possibility to improve a given route by replacing the two arcs $(i,j)$ and $(k,l)$ by the arcs $(i,k)$ and $(j,l)$; this implies the reversal of a sub-sequence between visits $j$ and $k$, $j$ and $k$ included.
\end{itemize}

\subsection{Iterative remove-repair heuristic}\label{sec:fast}

We also propose in this paper a fast heuristic called iterative removal/repair procedure (IRRP). This fast heuristic is used in several parts of the eILS. Algorithm \ref{algo:irrp} outlines the general heuristic structure. The main difference between IRRP and classic ILS is that it does not contain an intensification phase. The algorithm starts from an arbitrary solution (empty, partial or a complete one) and iteratively performs a removal/repair perturbation phase (See Section \ref{sec:removeRepaiOperator}). Once a new solution is constructed, it undergoes first an acceptance process to verify whether the incumbent solution is going to be updated. Subsequently, if an improvement in the objective function is achieved, $S_{best}$ is updated. Also, the value of $D_{max}$ is updated at the end of each iteration.

\begin{algorithm}[!ht]
\SetKwInOut{Input}{input}
\SetKwInOut{Output}{output}
\DontPrintSemicolon
\Input{Solution $S$.}
 $S_{Best} \leftarrow S$\;
 $S_{Incumb} \leftarrow S$\;
 $init(D_{max})$\;
  \While{$(!StopCondition)$}{
	$S_{Tmp} \leftarrow removalRepairPerturbation(S_{Incumb},D_{max})$ (See Section \ref{sec:removeRepaiOperator}) \;
	
	\lIf{$(AcceptSolution(S_{Tmp}))$}{
		$S_{Incumb} \leftarrow S_{Tmp}$
	}
	
  	\lIf{$(f(S_{Incumb}) < f(S_{Best}))$}{
  		$S_{Best} \leftarrow S_{Incumb}$
  	}
  	$Update(D_{max})$\;
 }
 \Return $S_{Best}$
\caption{{\sc Iterative removal/repair procedure}}\label{algo:irrp}
\end{algorithm}

\subsection{Iterated local search}\label{sec:ils}

The ILS is a metaheuristic scheme introduced by Louren{\c{c}}o et al. (2003) \cite{lourencco2003}. This approach is known  by its high potential, ease of implementation and few number of parameters to tune \cite{kramer2019}. A typical ILS algorithm is generally composed of three components: generation of an initial solution, perturbation phase, and local search procedure. The perturbation and local search procedures are iteratively applied to construct a new solution at each iteration. Instead of starting each time from scratch or the same base solution, they use the solution of the previous iteration as a starting point. The role of the perturbation phase is to prevent the metaheuristic from being trapped in local optima, whereas the local search aims at finding new local optimal solutions. The series of local optima produced by this process can be considered as a single chain of solutions followed by the ILS. Algorithm \ref{algo:ils} depicts the general scheme of the ILS. $D_{max}$ is called the perturbation degree and it is provided as a parameter to the perturbation procedure. It is initialized with a small value, and then incremented after each iteration without improvement. It is then reset to its initial value after each improvement. The stopping condition used in the ILS is $N$ iterations without improvement. The best solution is updated after each improvement of the total duration (line 7). It is noteworthy to mention that the fitness of partial solutions is computed as the sum of its total duration and the number of unscheduled tasks multiplied by a penalty. In our case, we set the penalty to $10^3$.

\begin{algorithm}[!ht]
\SetKwInOut{Input}{input}
\SetKwInOut{Output}{output}
\DontPrintSemicolon
\Input{Initial solution $S$.}
 $S \leftarrow Intensification(S)$ (See Section \ref{sec:intensification}) \;
 $S_{ILS} \leftarrow S$\;
 $D_{max} \leftarrow D_0$\;
 \While{$(! StopCondition)$}{
	$S \leftarrow Perturbation(S,D_{max})$ (See Section \ref{sec:perturbation})\;
  	$S \leftarrow Intensification(S)$ (See Section \ref{sec:intensification}) \;
  	\If{$(f(S) < f(S_{ILS}))$}{
  		$S_{ILS} \leftarrow S$\;
  	}
  	$Update(D_{max})$ 
 }
 \Return $S_{ILS}$
\caption{{\sc Iterated local search algorithm}}\label{algo:ils}
\end{algorithm}

An important issue with the ILS, is that by moving from one local optimum to another after each iteration, the exploration of the surrounding neighborhood of each solution is limited, which may cause the ILS to miss some good improving solutions. This aspect is pointed out by the proximate-optimality principle (POP) \cite{brandao2018} which suggests that good solutions are close to each other. To address this drawback, we further enhance the ILS by storing a set of elite solutions that are supplied by local optima found so far, and used later as a base solution for the ILS (See Section \ref{sec:elite}).

\subsubsection{Intensification phase}\label{sec:intensification}

A key feature of our approach is its reliance on an intensification phase that combines IRRP with a set of local search operators.

The procedure is a version of IRRP (see Section \ref{sec:fast}) that starts from a complete or a partial solution. Intensification is achieved by fixing $D_{max}$ to a relatively small value (see Section \ref{param:setting}), and performing no more than $N$ iterations. In this version of the IRRP, the solution provided int the input is maintained as the incumbent solution during all the process (Algorithm \ref{algo:irrp}, line 6), and it is only updated if the objective value is improved. A small and fixed value of 
$D_{max}$ (Algorithm \ref{algo:irrp}, line 8) allows the $IRRP$ to extensively explore the neighborhood of the current solution. During the removal repair perturbation phase, the aim is to improve the total duration of the incumbent solution. It is noteworthy to mention that we consider one removal operator rather than three (see Section \ref{sec:removeRepaiOperator}) in the IRRP used during intensification, which is the \textit{random removal} operator.

Algorithm \ref{algo:vnd} shows a pseudo code for the intensification procedure. The algorithm sequentially executes the operators one after another and iterates as long as there is at least one of the operators that succeeds in improving the current solution. After preliminary experimentation, $applySwap-sequence(S)$ only calls \textit{Swap-sequence(3,k)}, whereas $applySwap-relocate(S)$ explores all the combinations described earlier (see Section \ref{sec:lsOperators}).

\begin{algorithm}[!ht]
\DontPrintSemicolon
   \KwIn{\textit{Solution} $S$}
	$Imprv \leftarrow True$\;
	\While{$(Imprv = True)$}
 	{
 		$Imprv \leftarrow False$\;
  		\lIf{$(applyIRRP(S))$}{$Imprv \leftarrow True$}
 		\lIf{$(applyShift^1(S))$}{$Imprv \leftarrow True$}
  		\lIf{$(applyExchange^1(S))$}{$Imprv \leftarrow True$}
  		\lIf{$(apply2OptStar(S))$}{$Imprv \leftarrow True$}
  		\lIf{$(apply2Opt(S))$}{$Imprv \leftarrow True$}
  		\lIf{$(applyROpt(S))$}{$Imprv \leftarrow True$}
  		\lIf{$(applySwap-relocate(S))$}{$Imprv \leftarrow True$}
  		\lIf{$(applySwap-sequence(S))$}{$Imprv \leftarrow True$}
  	}
\Return $S$
\caption{{\sc Intensification procedure}}\label{algo:vnd}
\end{algorithm}

\subsubsection{Perturbation phase}\label{sec:perturbation}

Two perturbation approaches have been commonly used in the literature. The first approach is based on removal/repair procedures (see Section \ref{sec:removeRepaiOperator}), whereas the second one is based on local search operators. Hereafter, we discuss the implementation of these perturbation mechanisms in our method. 

This perturbation procedure is based on the \textit{Swap-sequence} operator described earlier (see Section \ref{sec:lsOperators}). A similar approach can be found in \cite{xie2017}. It applies a series of feasible but non-improving moves that deteriorate the fitness of the solution given as an input. Moreover, this procedure is designed such that the order in which the perturbation moves are performed is different from that used during the intensification phase. This is achieved in a similar way as in Brandao (2020) \cite{brandao2020}. First, a data structure (an array) is used to store the number of times each task is involved in local search moves performed during the intensification phase so far. Then, at the beginning of the perturbation phase, the array of counters is sorted in non-increasing order. For each task in this data structure, non-improving moves involving the sub-sequence that starts from this task are computed, and a random move is selected and performed. This process iterates over the list of tasks in a cyclic fashion, until the maximum number of moves is reached. The maximum number of moves is determined as a function of the perturbation degree parameter (see Section \ref{sec:sol}).

\subsection{Elite solutions}\label{sec:elite}

As explained in the previous sections, the ILS moves systematically from a local optimum to another after each iteration, without necessarily exploring all the neighborhood of each local optimum. To tackle this issue, we propose to store in a list of a maximum size of $N_{Pop}$ a set of good and relatively diversified solutions called \textit{elite set}. The basic idea of this approach is that instead of starting ILS from an empty solution, 
we rather use the ILS to continuously improve the solutions in the \textit{elite set}. Basically, the \textit{elite set} of solutions undergoes a series of updates that add new solutions of good quality and sufficiently diversified, and simultaneously discard unpromising ones. This approach shares several characteristics with evolutionary algorithms (\cite{prins2004}, \cite{vidal2013}), where a population of solutions evolves over several generations.

The evaluation of the solutions present in the \textit{elite set} is of crucial importance, because it allows the systematic discarding of unpromising solutions and maintains a high level of diversity. To achieve this purpose, we consider a so-called \textit{biased fitness}, which takes into consideration, in addition to the fitness of the solution, its contribution to the diversity of the \textit{elite set} \cite{prins2004}. 
The contribution of each solution to the diversity of the \textit{elite set} is computed as the average distance between the current solution and the $n_{Close}$ closest solutions in the \textit{elite set}. This parameter is fixed to a value of $20\%$ of the size of the population. Solutions are then ordered according to a weighted sum of their fitness and diversity contributions; and the best $N_{Pop}$ solutions are maintained, and the others are discarded.

\subsection{Generation of Initial Solutions}\label{sec:init}

This procedure is a version of IRRP that starts from an empty solution. The incumbent solution is always updated using the newly constructed solution in the previous iteration (Algorithm \ref{algo:irrp}, line 6). $D_{max}$ is set to its initial value which is equal to $D_0 = 3$. The value of $D_{max}$ is incremented by 1 after each iteration without improvement (Algorithm \ref{algo:irrp}, line 8), and reset to its initial value once a new best solution is found. The stopping condition of IRRP is $2N+K$, where $N$ is the number of tasks and $K$ is the number of vehicles. The evaluation criterion used in IRRP is the total duration (see Section \ref{sec:moveEval}).

\subsection{General flow}\label{sec:flow}

The algorithm starts with the initialization of the \textit{elite set}. $K$ solutions are constructed from scratch using IRRP heuristic (lines 2-6) (see Section \ref{sec:init}). After computing the \textit{biased fitness} of each solution (see Section \ref{sec:elite}), $N_{Pop}$ solutions are stored whereas the others are discarded (line 7).

As described in Section \ref{sec:perturbation}, a set $\mathcal{P}$ of four different versions of ILS is considered.
\begin{itemize}
\item $ILS_1$ : the perturbation is performed in a remove-repair fashion while using duration as moves evaluation criteria.
\item $ILS_2$ : the perturbation is performed in a remove-repair fashion while using travel cost as moves evaluation criteria.
\item $ILS_3$ : the perturbation is performed based on a local search procedure and moves evaluations is based on duration.
\item $ILS_4$ : the perturbation is performed based on a local search procedure and moves evaluations is based on travel cost.
\end{itemize}

Two phases are considered in the algorithm of the eILS. During phase one, only $ILS_1$ is used to generate new solutions. After $CV_1$ iterations without improving the best solution, eILS passes to phase 2 where $ILS_2,ILS_3$, and $ILS_4$ are sequentially called in this order, and each one is executed during $CV_2$ iterations. Once a new improving solution is found, eILS switches systematically to phase 1. This logic is presented by the procedure $selectILS(i,lstImpr, CV_1, CV_2)$ (line 12) in Algorithm \ref{algo:ils}. 
After the experimentation, the values of $CV_1$ and $CV_2$ are set respectively to values $45$ and $30$.
Regarding the general flow, the algorithm fetches at the beginning of each iteration a solution $S$ from the \textit{elite set} using binary tournament, then applies on it a major perturbation operation with a perturbation degree equal to $N/2$ (line 10), and provides it to a variant of $ILS$, which is chosen (line 12) 
based on the current iteration $i$, the iteration of the last improvement $lstImpr$, as well as $CV_1$ and $CV_2$.
The best solution found by the ILS is then added to the \textit{elite set} and triggers an update operation, during which at most $N_{Pop}$ solutions are retained in the \textit{elite set}, whereas duplicates and unpromising solutions are removed (line 12). This process is iterated $N_{Ils}$ times. At the end of the eILS, the best solution in terms of objective value present in the \textit{elite set} is returned (line 14).

\begin{algorithm}[!ht]
\SetKwInOut{Input}{input}
\SetKwInOut{Output}{output}
\DontPrintSemicolon
\Input{ILS versions $\mathcal{P}$, number of task $N$, convergence speed $CS_1$ and $CS_2$}
\Output{Best solution found so far $S_{Best}$.}
 $\mathcal{S} \leftarrow $
 $Pop \leftarrow \emptyset$\;
 \For{$i = 1$ \textbf{to} $K$}{
 	$S \leftarrow \emptyset$\;
 	$S_{IRRP} \leftarrow IRRP(S) \quad $ (see Section \ref{sec:fast})\;
 	$Pop \leftarrow Pop \cup \{S_{IRRP}\}$
 }
 $Pop \leftarrow initPopulation(Pop) \quad $ (see Section \ref{sec:elite})\;
 $lstImpr \leftarrow 0$\;
	\For{$(i = 1$ \textbf{to} $N_{Ils})$}{
	  $S \leftarrow BinaryTournement(Pop)$\;
	  $S \leftarrow removalRepairPerturbation(S,N/2)$ (see Section \ref{sec:removeRepaiOperator}) \;
	  $ILS \leftarrow selectILS(i,lstImpr,\mathcal{P}, CS_1, CS_2)$\;
 	  $S_{ILS} \leftarrow ILS(S) \quad $ (see Section \ref{sec:ils})\;
	  $Pop \leftarrow updatePopulation(Pop,S_{ILS}) \quad $ (see Section \ref{sec:elite})\;
	  \If{$(fitness(S_{ILS}) < fitness(S_{Best}))$}{
	  	$S_{Best} \leftarrow S_{ILS}$\;
	  	$lstImpr \leftarrow i$\;
	  }
 }
 $S_{Best} \leftarrow Top(Pop)$\;
\Return $S_{Best}$
\caption{{\sc Resolution algorithm}}\label{algo:ils}
\end{algorithm}

\subsection{Move evaluation and feasibility check}\label{sec:moveEval}

The TRSP is a rich vehicle routing problem \cite{caceres2014}, with several characteristics and constraints, such as time windows, renewable and non-renewable resources, multi-depot, and site-dependent considerations. The difficulty of the problem is further increased when dealing with duration minimization rather than travel costs minimization. Hence, the need for a constant-time cost evaluation and feasibility checks, for both, repair and local search moves, is crucial for the overall performance of our method.

Before proceeding further, let us consider $\sigma$ as a sequence of arbitrary visits of nodes, which can be task locations, home depots, or the central depot. We denote the node at the $i^{th}$ position of $\sigma$ by $\sigma^i$.

We also define the concatenation operator of two or more sub-sequences $\sigma_1$ and $\sigma_2$ as :

\begin{equation}
\sigma = \sigma_1 \oplus \sigma_2
\end{equation}

Let $\Delta(\sigma)$, $\Lambda^r(\sigma)$, $\Lambda^n(\sigma)$ and $\Psi(\sigma)$ be duration of sequence $\sigma$, accumulated renewable resources of sequence $\sigma$, accumulated non-renewable resources of sequence $\sigma$ and the accumulated skills needed by tasks in sequence $\sigma$, respectively. We propose in the following the formulas used to perform feasibility checks and moves evaluation in a constant time.

\paragraph{\textbf{Renewable resources}}

For each task at position $i$ in the sequence $\sigma$, we record the number of times a given type of tool $t \in T$ is required by the tasks from the start of $\sigma$ to position $i$, or until the central depot in the case where it is visited before position $i$.

The following equations hold.

\begin{center}
$\Lambda^r_t(\sigma^i) =
\left\{
	\begin{array}{l}
		\Lambda^r_t(\sigma^{i-1}) +  b_{\sigma^i,t} \quad if \; \sigma^k \neq 0 \; \forall k \leq i-1\\
		\Lambda^r_t(\sigma^{i-1}) \quad otherwise
	\end{array}
\right.$
\end{center}

\begin{center}
$\Lambda^r_t(\sigma)$ = $\Lambda^r_t(\sigma^{|\sigma|})$.
\end{center}

\begin{center}
$\Lambda^r_t(\sigma) =
\left\{
	\begin{array}{l}
		\Lambda^r_t(\sigma_1) +  \Lambda^r_t(\sigma_2) \quad if \; 0 \notin \sigma_1 \\
		\Lambda^r_t(\sigma_1) \quad otherwise
	\end{array}
\right.$
\end{center}

Let $k \in \mathcal{K}$ be the vehicle performing sequence $\sigma$. $\sigma$ is unfeasible \textit{if} $ \exists t \in \{1,\ldots,|T|\}$ where $w^k_t = 0$ and $\Lambda^r_t(\sigma) > 0$.

\paragraph{\textbf{Non-renewable resources}}

For each task at position $i$ in the sequence $\sigma$, we record the number of times a given type of spare part $p \in P$ is required by the tasks from the start of $\sigma$ to position $i$, or until the central depot in the case where it is visited before position $i$.

The following equations hold.

\begin{center}
$\Lambda^n_p(\sigma^i) =
\left\{
	\begin{array}{l}
		\Lambda^n_p(\sigma^{i-1}) +  d_{\sigma^i,p} \quad if \; \sigma^k \neq 0 \; \forall k \leq i-1\\
		\Lambda^n_p(\sigma^{i-1}) \quad otherwise
	\end{array}
\right.$
\end{center}

\begin{center}
$\Lambda^n_p(\sigma)$ = $\Lambda^n_p(\sigma^{|\sigma|})$.
\end{center}

\begin{center}
$\Lambda^n_p(\sigma) =
\left\{
	\begin{array}{l}
		\Lambda^n_p(\sigma_1) +  \Lambda^n_p(\sigma_2) \quad if \; 0 \notin \sigma_1 \\
		\Lambda^n_p(\sigma_1) \quad otherwise
	\end{array}
\right.$
\end{center}

Let $k \in \mathcal{K}$ be the vehicle performing sequence $\sigma$. $\sigma$ is unfeasible \textit{if} $ \exists p \in \{1,\ldots,|P|\}$ where $v^k_p > \Lambda^n_p(\sigma)$. 

\paragraph{\textbf{Skills}}

For each task in a sequence $\sigma$, we record the number of times a given skill is required by the tasks since the start of the sequence. For each type of skill $q \in Q$, the following equations hold. $\Psi_q(\sigma^i)$ = $\Psi_q(\sigma^{i-1})$ + $a_{\sigma^i,q}$. $\Psi_q(\sigma)$ =  $\Psi_q(\sigma^{|\sigma|})$.
$\Psi_q(\sigma)$ = $\Psi_q(\sigma_1)$ + $\Psi_q(\sigma_2)$. Let $k \in \mathcal{K}$ be the technician performing sequence $\sigma$. $\sigma$ is unfeasible \textit{if} $ \exists q \in \{1,\ldots,|Q|\}$ where  $y^k_q = 0$ and $\Psi_q(\sigma) > 0$.

\paragraph{\textbf{Time window feasibility}}

To perform time window feasibility checks in a constant time, we adopted the approach proposed by Kindervarter and Savelsbergh (2018) \cite{kindervater2018}, where the authors proposed to compute the \textit{Forward Time Slack} $FTS_i$ at the $i^{th}$ position of $\sigma$, indicating how much time delay is possible at this position while maintaining time windows feasibility of the subsequent visits. Let $WT_i$ and {\color{red}{$h^\sigma_i$}} be the waiting time and the service starting time at the $i^{th}$ position of $\sigma$, respectively. We define the total waiting time $TWT_{ij}$ between $\sigma_i$ and $\sigma_j$, $i \leq j$ as follows: $TWT^\sigma_{ij} = \sum_{k = i+1}^j WT_k$.

The forward time slack at the $i^{th}$ position of $\sigma$ is defined as :  $FTS^\sigma_i = \min\limits_{i \leq k \leq |\sigma|}\{ TWT^\sigma_{ik}  + l_{\sigma_k} - h^\sigma_k \}$, where $h^\sigma_k$ is the service starting time of the visit at position $k$ of $\sigma$. For convenience, we denote the FTS at the starting position as $FTS^{\sigma}$. The insertion of a task $r \in R$ in the $i^{th}$ position of $\sigma$ is feasible if : $s_r < l_r  \quad and\quad Shift_i < FTS^\sigma_i$, where $s_r$ is calculated as follows: $s_r = max\{h^\sigma_{i-1} + \delta_{\sigma_{i-1}} + t_{\sigma_{i-1},r}, e_r\}$ and $Shift_i = \tilde{h}^{\sigma}_i - h^\sigma_i$, where $\tilde{h}^{\sigma}_i = max\{s_r + \delta_r +  t_{r,\sigma_{i}}, e_{\sigma_i}\}$.

\paragraph{\textbf{Movement evaluation}}
The objective function of the TRSP is to minimize the total duration of the routes. This also includes the possibility of delaying the departure of vehicles from the depots to minimize the total waiting time. We proceed in a similar manner as in Savelsbergh. (1992) \cite{savelsbergh1992}. Let us first define the maximum delay of the service starting time at the $i^{th}$ position of $\sigma$ without violating time windows at subsequent visits or causing any delays at the arrival depot as the \textit{Passive Time Slack} (PTS) of $\sigma$, and it is denoted by $PTS_i^\sigma$. This is calculated as follows : $PTS_i^\sigma = \{FTS^\sigma_i,TWT^\sigma_{i|\sigma|}\}$. For convenience, we denote the PTS at the starting position as $PTS^{\sigma}$.

The duration of sequence $\sigma$ is computed as 

$\Delta(\sigma) = h^\sigma_{|\sigma|} + \delta_{\sigma_{|\sigma|}} - h^\sigma_1 - PTS^\sigma$.

To compute the duration of the concatenated sequences, we compute the new PTS and the earliest arrival at the final position of the new sequence. For this purpose, we first define the \textit{allowed backward shift} of a given sequence $\sigma$, at a given position $i$, as the maximum gain in duration at the final depot yielded by the backward shift of the service starting time at position $i$ of $\sigma$, assuming of course that the service can start in an earlier date. The \textit{allowed backward shift} is denoted by $BS_i^{\sigma}$ and it is computed as follows: $BS^{\sigma}_i=min\{h^\sigma_i - e_{\sigma_i},BS^{\sigma_{i+1}}\}$. For convenience, we note the BS at the first position of $\sigma$ by $BS^\sigma$.

Without loss of generality, we consider the concatenation of three sub-sequences into a single sequence: $\sigma = \sigma_1 \oplus \sigma_2 \oplus \sigma_3$. Let $Shift^{\sigma_2}$ and $Shift^{\sigma_3}$ be, respectively, the shifts at the first elements of $\sigma_2$ and $\sigma_3$ after concatenation. We also denote by $WT^{\sigma_2}$ and $WT^{\sigma_3}$ the new waiting times at the first elements of $\sigma_2$ and $\sigma_3$ after concatenation.

Depending on the value of $Shift^{\sigma_3}$, either positive or negative, we propose the following formulas to compute the duration of $\sigma$.\\

If $Shift^{\sigma_3} \geq 0$ :

\begin{itemize}
\item $h^\sigma_{|\sigma|} = h^{\sigma_3}_{|\sigma_3|} + max(0,Shift^{\sigma_3} - PTS^{\sigma_3})$
\item $PTS^\sigma = min\{FTS^{\sigma_1}, min\{ TWT^{\sigma_1} + WT^{\sigma_2} + max\{0 , FTS^{\sigma_2} - Shift^{\sigma_2}\} , TWT^{\sigma_1} + WT^{\sigma_2} + max\{0 , TWT^{\sigma_2} - Shift^{\sigma_2}\} + WT^{\sigma_3} + max(0,PTS^{\sigma_3} - Shift^{\sigma_3})\}\}$
\end{itemize}

If $Shift^{\sigma_3} < 0$ : 

\begin{itemize}
\item $h^\sigma_{|\sigma|} = h^{\sigma_3}_{|\sigma_3|} - min(-Shift^{\sigma_3}, BS^{\sigma_3})$
\item $PTS^\sigma = min\{FTS^{\sigma_1}, min \{TWT^{\sigma_1} + WT^{\sigma_2} + max\{0 , FTS^{\sigma_2} - Shift^{\sigma_2}\} 
, TWT^{\sigma_1} + WT^{\sigma_2} + max\{0 , TWT^{\sigma_2} - Shift^{\sigma_2}\} + WT^{\sigma_3} + max(0,PTS^{\sigma_3}-Shift^{\sigma_3} - BS^{\sigma_3})\}\}$
\end{itemize}

\subsection{Speed-up techniques}

Local search operators and the best insertion algorithm are the most time-consuming components of the proposed scheme. We propose in the following two speed-up techniques used to reduce the computational burden.

\subsubsection{Parallel best insertion algorithm}

An efficient method to implement the best insertion algorithm is to consider each route separately. The basic idea is to compute the best feasible insertion of the unscheduled tasks for each route. Then, those feasible insertions of all routes are stored in a heap structure that we call a HEAP. The insertion process is performed as follows. It starts by selecting the best move (if any) from HEAP. If the task has not yet been inserted, the insertion is performed and the task is marked as fulfilled and removed from the list of unscheduled tasks. The algorithm systematically computes feasible insertions of the remaining unscheduled tasks while considering only the last modified route. If feasible moves are found, then the best move is selected and pushed to HEAP. This process is iterated until HEAP becomes empty, i.e. no route has feasible insertions in HEAP.

\subsubsection{Nearest predecessors}

Many irrelevant movement evaluations are performed during local search calls. Experimentation showed that most improvements are achieved by local search movements involving the nearest neighbors. To take advantage of this observation, we propose to compute a list of the nearest predecessors for each task. Given an arbitrary task $i$, and for each predecessor $j$, and assuming that arc $(j,i)$ is feasible, the distance is computed as follows : 

\begin{equation}
dist(j,i) = max\{max(0,e_i - l_j), t_{ji}\}.
\end{equation}

The predecessors of $i$ are sorted in non-decreasing order according to distance and only the first $\chi$ predecessors are considered during local search movements. 
The value of $\chi$ was fixed to a value equal to $30$ after experimentation.

The nearest predecessors lists are mainly used by inter-route local search operators. The $2-opt*$ operator, for example, where arcs $(i,j)$ and $(k,l)$ are interchanged, it starts by selecting a task $j$, and then picks a task $l$ among the $\chi$ nearest predecessors. $l$ should be scheduled on a route different from that of task $j$. Move evaluation in $Swap-Relocate$ and $Swap-Sequence$ are carried out in the same fashion.

\section{Computational results}\label{sec:comput}
We present in this section the computational tests carried out to assess the performance of the proposed eILS. Our algorithm is coded in C++ and runs on a PC with an Intel Core i7 2.6GHz processor and 16 GB RAM. The results of the eILS are compared with those of parallel ALNS (pALNS) algorithm found in \cite{pillac2013}.

\subsection{Benchmark instances}
Benchmark instances for TRSP are derived from 56 instances of Solomon benchmark for VRPTW \cite{solomon1987}. All instances have 100 tasks, that are either randomly distributed (R), in clusters (C) or a mix patterns (RC). Each class can be grouped into two sub-classes: (1) instances with short time windows or (2) large time windows (i.e. C1, C2, R1, R2, RC1, RC2). For each instance, 25 home depots are added, each associated with a technician and initial inventory of tools and spare parts. Each technician is associated with a set of skills (five types of skills are considered) and each task is associated with requirements in terms of skills, tools, and spare parts. The depot present in the original instances of VRPTW is considered the central depot used for replenishment. As indicated in the problem definition, the central depot has unlimited inventory, and each technician can carry as many tools and spare parts as needed for their journey without any capacity limitations. 

\subsection{Parameter settings}\label{param:setting}

In this section, we investigate the impact of several parameters of the eILS on the overall performance. Three parameters were determined: number of tasks to remove $D_{max}$ at each iteration in the $IRRP$ during the intensification phase, size of the \textit{Elite set} $N_{pop}$, and number of iterations $N_{ils}$ of eILS. The values for each parameter are listed in Table (\ref{tab:params2values}).

{\centering
\begin{table}[!ht]
\setlength\extrarowheight{2.5pt}
\caption{Parameter settings}
\label{tab:params2values}
\footnotesize
\begin{tabular}{llc}
\toprule
Parameter & Description & Range\\
\midrule
$d_{max}$& Perturbation parameter in $IRRP$ & $[5,10,15]$\\
\midrule
$N_{ils}$& Number of Iterations & $[450,500,550,600]$\\
\midrule
$N_{pop}$& Size of elite set & $[10,15,20]$\\
\bottomrule
\end{tabular}
\end{table}
}

\begin{figure}
\begin{minipage}{.5\linewidth}
\begin{center}
\includegraphics[scale=0.5]{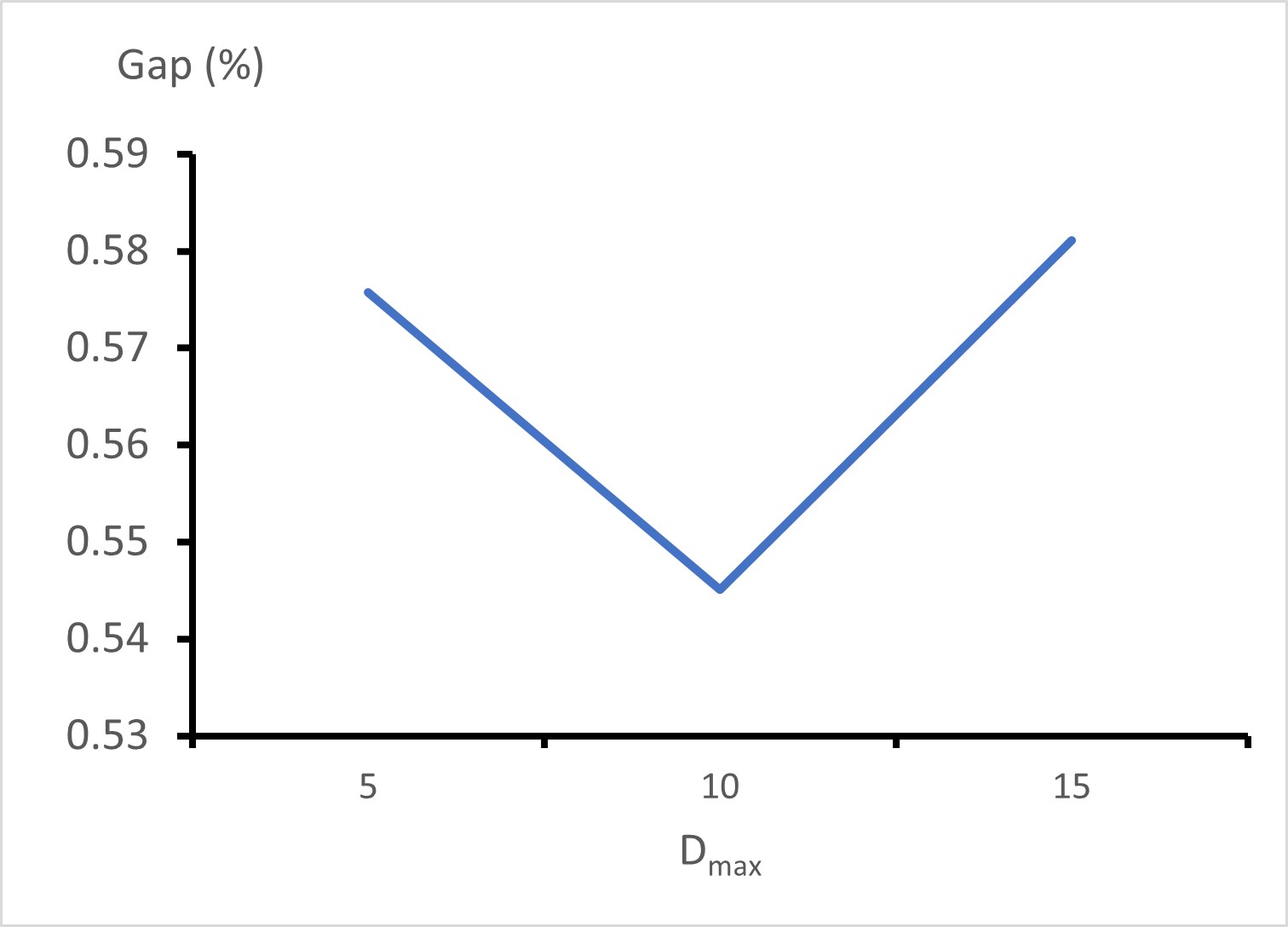}
\end{center}
\caption{Impact of $D_{max}$ in IRRP on overall gap}
\label{fig:impactDmaxGap}
\end{minipage}
\begin{minipage}{.5\linewidth}
\begin{center}
\includegraphics[scale=0.5]{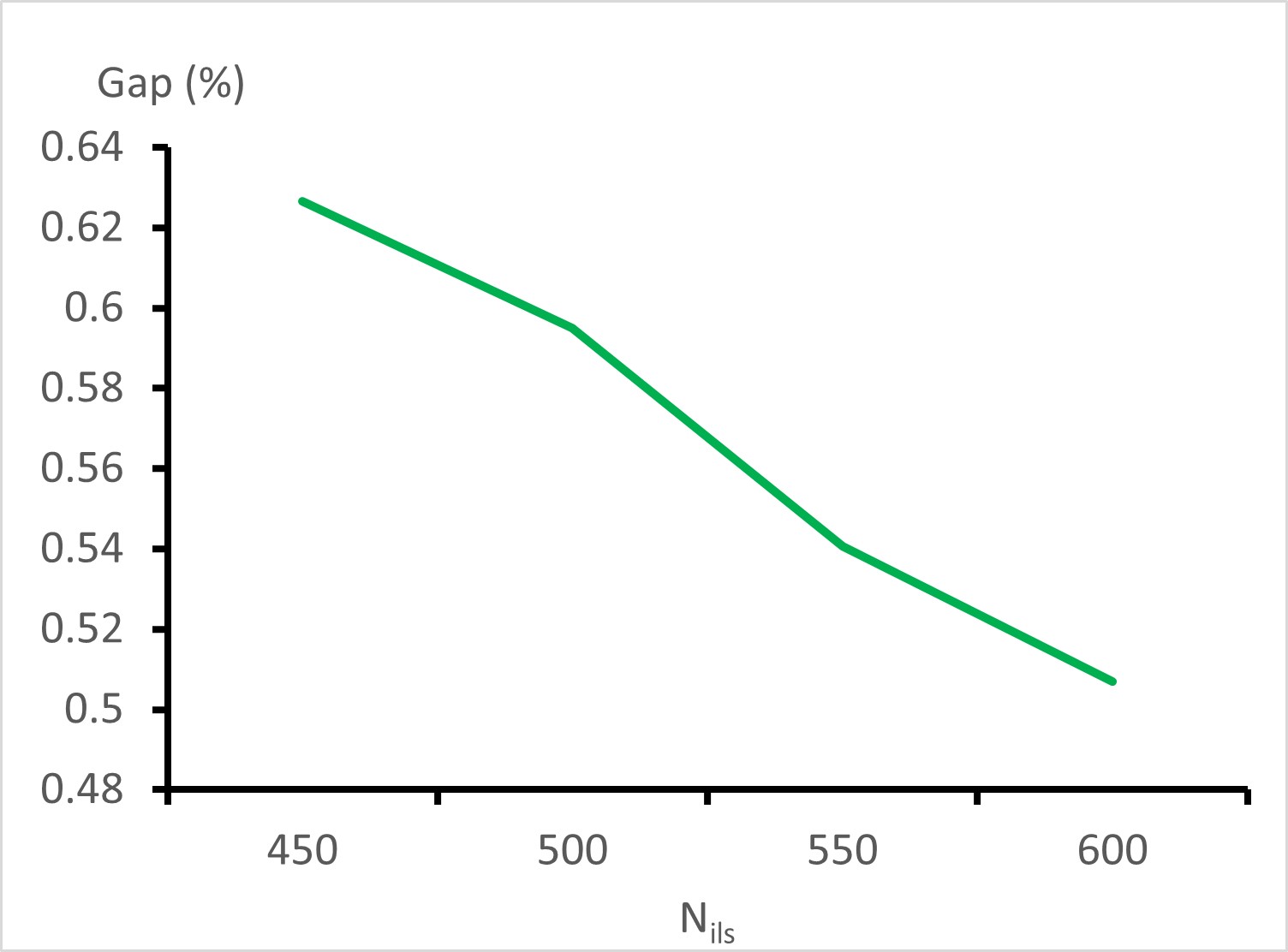}
\end{center}
\caption{Impact of the number of iterations on overall gap}
\label{fig:impactNilsGap}
\end{minipage}
\end{figure}

\begin{figure}
\begin{center}
\includegraphics[scale=0.5]{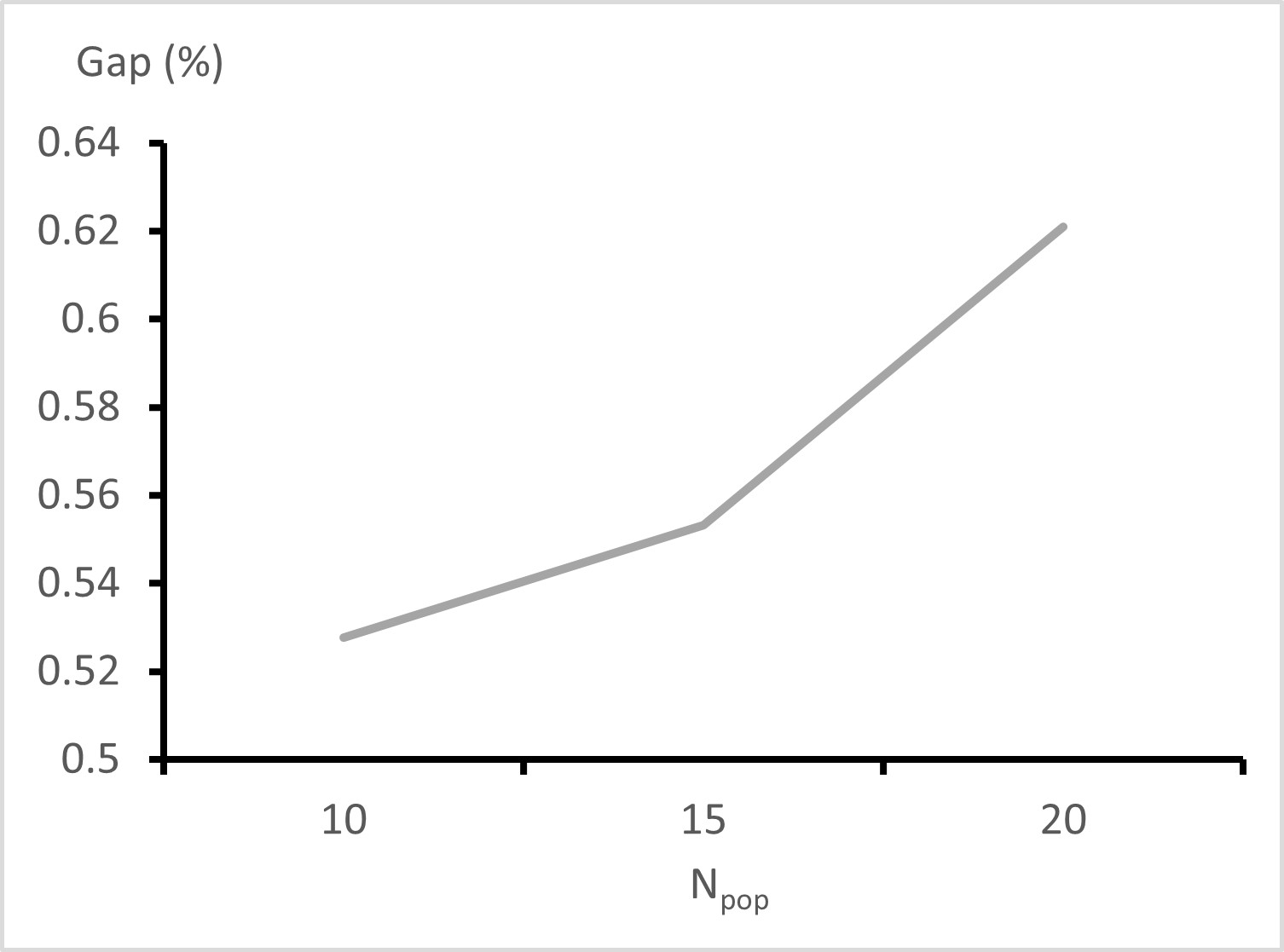}
\end{center}
\caption{Impact of the population size on overall gap}
\label{fig:impactPopSizeGap}
\end{figure}

\begin{figure}
\begin{center}
\includegraphics[scale=0.5]{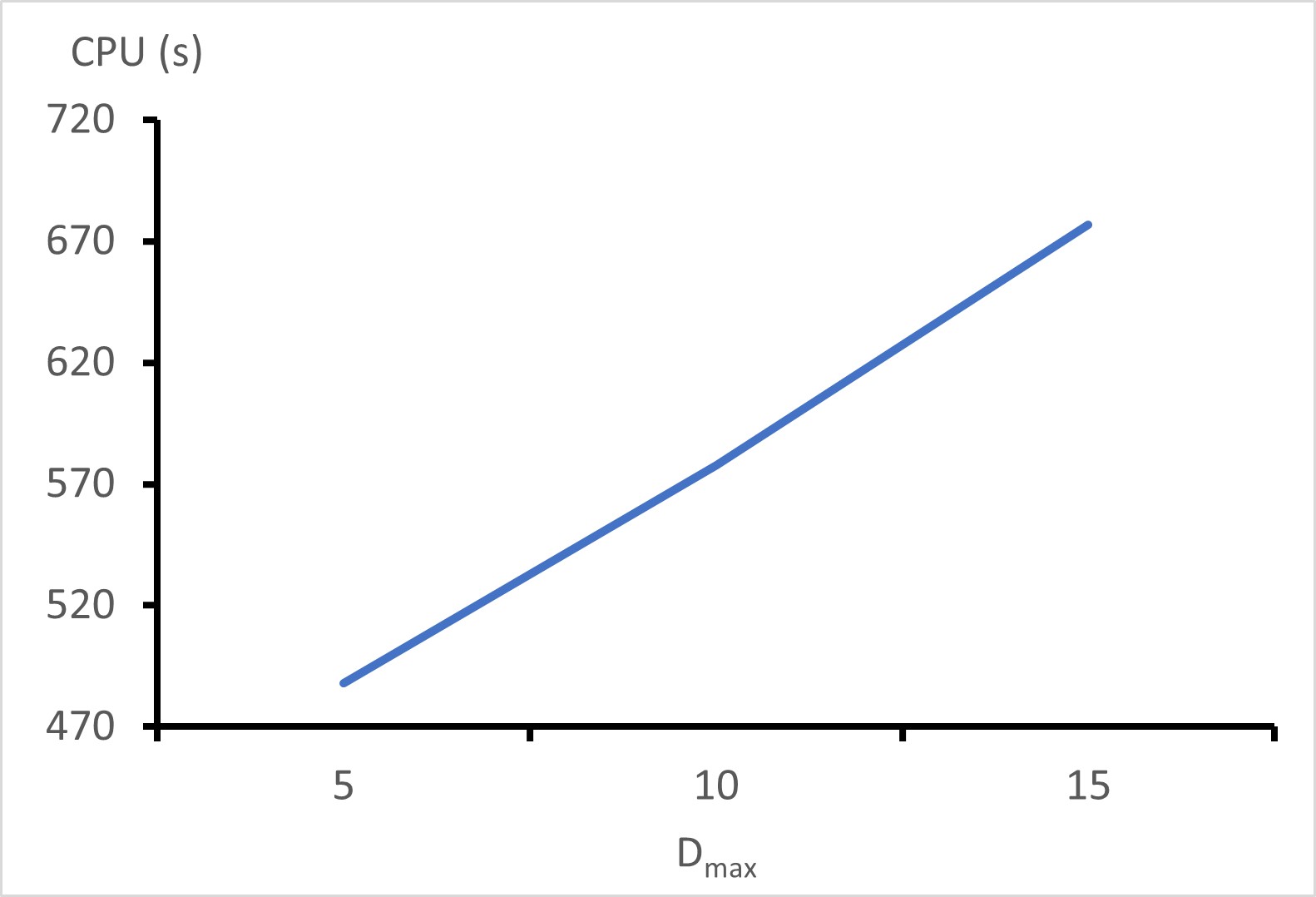}
\end{center}
\caption{Impact of the number of iterations on overall computational time}
\label{fig:impactDmaxCPU}
\end{figure}

\begin{figure}
\begin{minipage}{.5\linewidth}
\begin{center}
\includegraphics[scale=0.5]{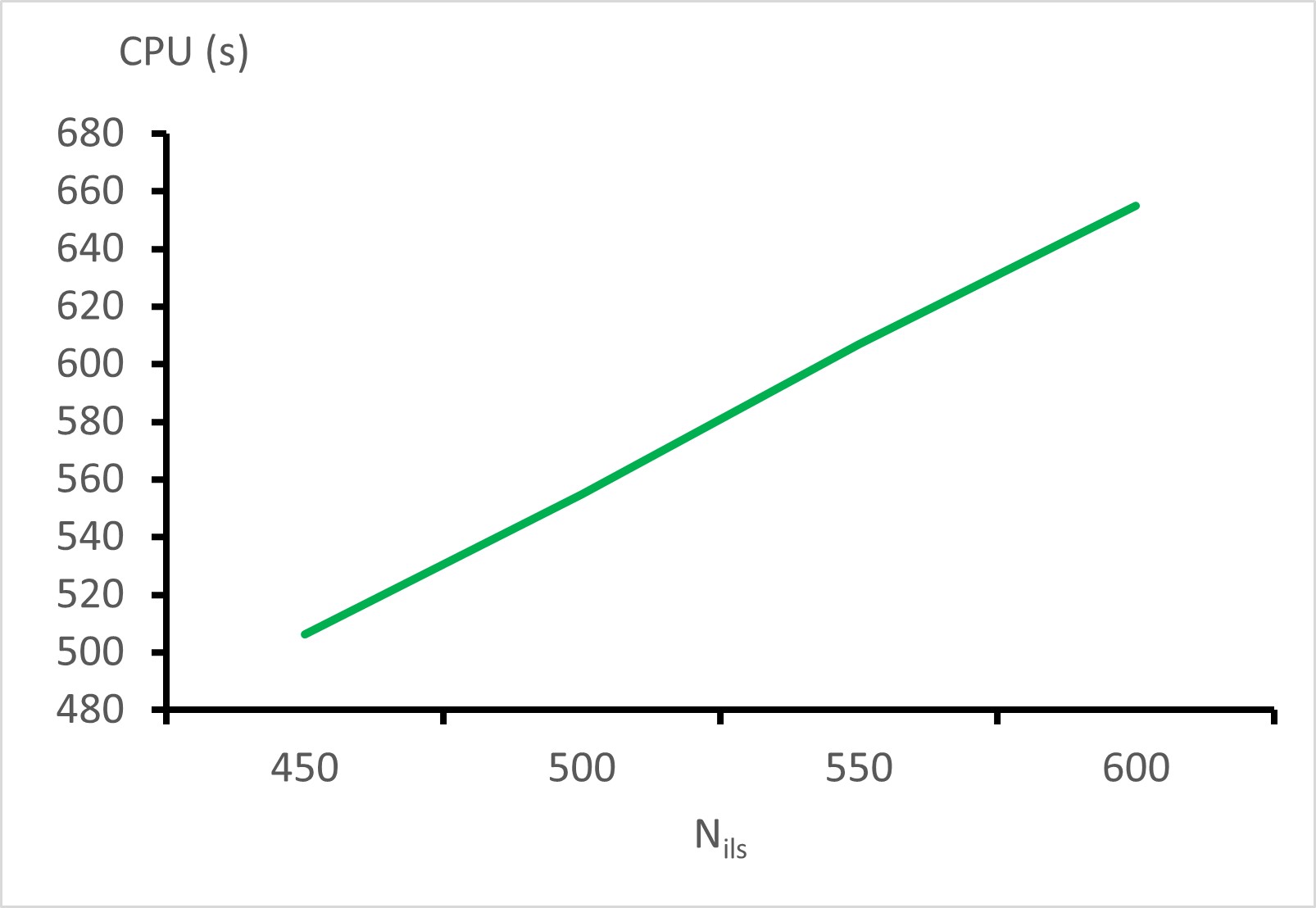}
\end{center}
\caption{Impact of the number of iterations on overall computational time}
\label{fig:impactNilsCPU}
\end{minipage}
\begin{minipage}{.5\linewidth}
\begin{center}
\includegraphics[scale=0.5]{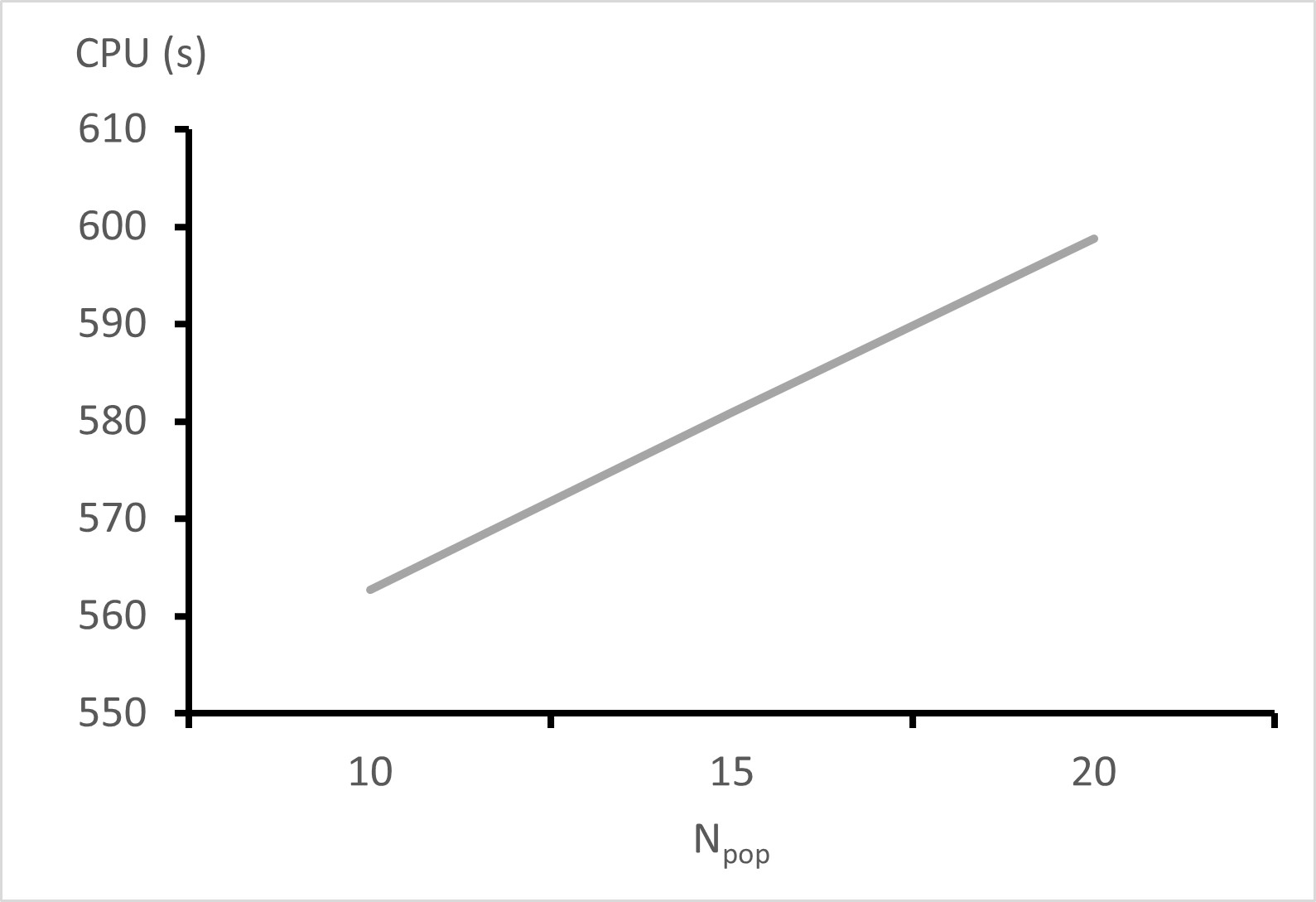}
\end{center}
\caption{Impact of the population size on overall computational time}
\label{fig:impactPopSizeCPU}
\end{minipage}
\end{figure}

Twelve instances were arbitrarily selected for this purpose, and each instance was executed 10 times for each combination of parameter values, that is, $144*10$ runs for each instance. Figures 
\ref{fig:impactDmaxGap},\ref{fig:impactNilsGap} and \ref{fig:impactPopSizeGap} show the aggregated overall gaps for each combination of the three parameters, whereas figures \ref{fig:impactDmaxCPU},\ref{fig:impactNilsCPU} and \ref{fig:impactPopSizeCPU} show the overall computational times variations according to the values of the same parameters.

Based on this experiment, the selected value of each parameter are listed in Table \ref{tab:params2bestvalues}.

{\centering
\begin{table}[!ht]
\setlength\extrarowheight{2.5pt}
\centering
\caption{Parameter settings for the eILS}
\label{tab:params2bestvalues}
\begin{tabular}{ccccc}
\toprule
Parameters &  $d_{max}$ & $N_{ils}$ & $N_{pop}$\\
\midrule
Value & $10$ & $600$ & $10$\\
\bottomrule
\end{tabular}
\end{table}
}

\subsection{Sensitivity analysis}

In this section, we present a study of the sensitivity analysis of several components of the eILS. We start by investigating the contribution of each perturbation mechanism, then, we focus on two components of the eILS, which are the related-removal operator and \textit{Swap-Sepquence} local search operator.

\subsubsection{Perturbation mechanisms}

We focus in this section on the contribution of each of the four perturbation mechanisms (see Section \ref{sec:flow}). We ran our algorithm 10 times while considering several configurations of the eILS. These configurations are the following:

\begin{itemize}
\item $Conf_1$ = $ILS1$
\item $Conf_2$ = $ILS1 + ILS2$
\item $Conf_3$ = $ILS1 + ILS2 + ILS3$
\item $Conf_4$ = $ILS1 + ILS2 + ILS3 + ILS4$ = eILS
\end{itemize}

We provide the following performance indicators for the methods:

\begin{itemize}
\item $CPU$ : the average computational times of the 10 runs per sub-class. In this section, we take as a reference the computational times of $Conf_1$.
\item $GAP$ : the gap between the best objective value of a given method and the best-know solution per sub-class, and it is calculated as follows:
\begin{equation}
GAP = \frac{C_{best}-C^M_{best}}{C_{best}} \times 100\label{expr:gap}
\end{equation}
where $C_{best}$ is the best-known solution and $C^M_{best}$ is the best objective value of the 10 runs of either pALNS or eILS.
\item $DEV$ : the deviation of the average best results from the best objective values per sub-class.
\begin{equation}
DEV = \frac{C^M_{best}-AVG^M}{C^M_{best}} \times 100\label{expr:dev}
\end{equation}
where $AVG^M$ is the average objective value of the ten runs given by the eILS or pALNS.
\end{itemize}

\begin{table}[!ht]
\setlength\extrarowheight{2.5pt}
\centering
\caption{Sensitivity analysis of perturbation mechanisms}
\label{tab:perturbMechanism}
\begin{tabular}{lcccc}
\toprule
Parameters & $Conf_1$ &  $Conf_2$  & $Conf_3$ & $Conf_4$\\
\midrule
$GAP (\%)$ & $0.243$ & $0.21$ & $0.198$ & $0.19$\\
\midrule
$DEV (\%)$& $0.707$ & $0.673$ & $0.7$ & $0.684$ \\
\midrule
$CPU$& $1$ & $1$ & $1.14$ & $1.02$ \\
\bottomrule
\end{tabular}
\end{table}

Table \ref{tab:perturbMechanism} provides performance measures for the configurations presented in this section. This clearly shows the contribution of each ILS version (perturbation mechanism) to the global performance of eILS. Adding $ILS2$ to $ILS1$ ($Conf_2$) has clearly improved the overall gap (row 2) from $0.243\%$ to $0.21\%$, the deviation from $0.707\%$ to $0.673\%$, 
Interestingly, these substantial improvements have been achieved while maintaining similar computational times.
Adding $ILS3$ in $Conf_3$ has also improved the the overall gap from $0.21\%$ to $0.198\%$, although a slight deterioration has been observed on the deviation from the objective value (row 3) and a substantial increase in computational times by a factor of $1.14$.
Finally, the full scheme of eILS has achieved the best global performance in terms of the overall gap with an associated value of $0.19\%$. The overall deviation has slightly deteriorated from $0.673\%$ to $0.684\%$. The computational times have slightly increased by a factor of $1.02$ compared to $Conf_1$.

\subsubsection{Components sensitivity analysis}

In this section, we suggest to investigate the contribution of some of eILS components to the overall performance. We focus our attention on two components, namely, \textit{related-sequence} removal, referred to by $SeqRem$, and \textit{swap-sequence} local search operator, referred to by $SwapSeq$. 
A similar approach of \textit{related-sequence} removal operator but more complex can be found in \cite{christiaens2020}.

\begin{table}[!ht]
\setlength\extrarowheight{2.5pt}
\centering
\caption{Sensitivity analysis for some components of the eILS}
\label{tab:sensAnalysis}
\begin{tabular}{cccc}
\toprule
Parameters & no $SwapSeq$ &  no $SeqRem$ & eILS \\
\midrule
$GAP (\%)$ & $0.210$ & $0.192$ & $0.179$\\
\midrule
$DEV$ $(\%)$ & $0.810$ & $0.607$ & $0.629$\\
\midrule
$CPU$& $0.84$ & $0.88$ & $1$ \\
\bottomrule
\end{tabular}
\end{table}

Table \ref{tab:sensAnalysis} shows a comparison of the three versions of eILS. The first version does not include the related sequence removal operator (column 1), whereas the second version does not include the swap sequence operator (column 2). The last column of Table \ref{tab:sensAnalysis} shows the results of eILS, including both components. The results clearly show the contribution of both components to the overall results of the eILS. The absence of \textit{related-sequence} removal (column 2) substantially deteriorates the quality of solutions obtained by the eILS. More precisely, the percentage gap ($GAP$) evolves from $0.179\%$ to $0.192\%$. However, the computational times have substantially decreased by a factor of $0.88$ compared to eILS.

The same behavior has been observed when the \textit{swap-sequence} local search operator is discard. The percentage gap goes from $0.179\%$ to $0.210\%$ whereas the percentage deviation has achieved $0.810\%$ compared to $0.629\%$ of the eILS. However, the computational times have substantially decreased by a factor of $0.84$ compared to the whole scheme of eILS.

\subsection{Computational results}
We conduct experiments to assess the performance of our method. We compare our method with the pALNS presented in \cite{pillac2013}. The pALNS was implemented using Java 7 and Gurobi 4.60 on an Ubuntu 11.10 64-bit machine, with an Intel i7 860 processor (4×2.8GHz) and 6GB of RAM, using K = 8 subprocesses. To guarantee a fair comparison between the two methods, we adopt the same protocol used in \cite{pillac2013}, that is, we perform ten random runs of the eILS on each instance tested, and we report the best objective value, the average objective value, and the average computational times. 

Table \ref{tab:pALNSSC} shows a comparison between the two methods.

We recorded the deviation from the average ($DEV$) computed using Eq. (\ref{expr:dev}), the computational times ($CPU$) and the gap to the best solution ($GAP$) computed using Eq. (\ref{expr:gap}). The computational times and deviation from the average of the pALNS can be found in \cite{pillac2013}. Because pALNS is a parallel approach, the authors in \cite{pillac2013} reported two computational times: the computational times of the parallel ALNS $cpu_1$ and the computational times of post-optimization phase (route recombination) $cpu_2$. Hence, we estimate the computational times of a sequential version of pALNS as $CPU_{pALNS} = 8 \times cpu_1 + cpu_2$.

{\tiny
\begin{table}[!ht]
\setlength\extrarowheight{2.5pt}
\centering
\caption{Summary of results of eILS and pALNS}
\label{tab:pALNSSC}
\begin{tabular}{l|ccc|ccc}
\toprule
\multirow{2}{*}{Class} & \multicolumn{3}{c|}{pALNS} & \multicolumn{3}{c}{eILS}\\
\cmidrule(lr){2-4}\cmidrule(lr){5-7} &
$CPU$ $(s)$ & $GAP$ $(\%)$ & $DEV$ $(\%)$ & $CPU$ $(s)$ & $GAP$ $(\%)$ & $DEV$ $(\%)$ \\
\midrule
$C1$&$580.9$&$0.060$&$0.23$&$329.97$&$0.054$&$0.291$\\
\midrule
$C2$&$246$&$0.474$&$0.42$&$414.88$&$0$&$0.112$\\
\midrule
$R1$&$731.4$&$0.094$&$0.82$&$343.87$&$0.351$&$0.886$\\
\midrule
$R2$&$290.1$&$1.5$&$1.46$&$382.03$&$0.065$&$0.818$\\
\midrule
$RC1$&$409$&$0.104$&$0.68$&$327.41$&$0.609$&$0.985$\\
\midrule
$RC2$&$238.8$&$1.466$&$1.43$&$369.17$&$0$&$1.111$\\
\bottomrule
$Mean$&$434.9$&$0.617$&$0.86$&$360.54$&$0.184$&$0.713$\\
\bottomrule
\end{tabular}
\end{table}
}

The results in Table \ref{tab:pALNSSC} show no clear dominance between the two methods. However, we notice that for sub-classes with tight windows (R1, and RC1), pALNS succeeds in outperforming eILS, since it achieves a percentage gap of $0.094\%$ and $0.104\%$ respectively, against $0.351\%$, and $0.609\%$. This is mainly justified by the use of a powerful post-optimization approach based on a set covering formulation used in \cite{pillac2013}. Nevertheless, eILS improves the best know solutions of several instances of these classes. In contrast, for subclasses C1, C2, R2, and RC2, which are characterized by relatively large time windows, eILS outperforms pALNS in terms of percentage gaps and deviation from the average. More precisely, eILS reaches percentage gaps of $0.54\%$, $0\%$, $0.013\%$, and $0.062\%$ on these subclasses, against the percentage gaps of $0.060\%$, $0.522\%$, $1.512\%$, and $1.527\%$ achieved by the pALNS.

It is noteworthy to mention that the computational times of the eILS are much higher on subclasses with large time windows. This is mainly owing to the extensive use of local search operators, especially because these type of instances have room for improving solution costs, whereas for instances with tight time windows, the contribution of local search operators is limited to inter-route operators. As a consequence, eILS succeeds in improving almost all the instances of C2, RC2 and RC2, achieving a total number of 34 new best-known solutions (see \ref{app:detailed}).

Globally, although the eILS fails to improve the best-know solution for a number of instances in classes C1, R1 and RC1, it achieves a better overall gap to best, which is equal to $0.184\%$, which is more that three times lower than that of pALNS $(0.617\%)$. Regarding computational times, eILS has an overall CPU time equal to $360.54s$, against $434.9s$ for pALNS. 

\section{Conclusion and perspective}\label{sec:concl}

In this paper, we addressed a variant of the workforce scheduling problem called TRSP. This variant incorporates several state-of-the-art constraints such as \textit{time windows}, \textit{multi-depot}, \textit{site-dependent}, \textit{capacity constraints}, etc. We proposed in this paper an enhanced version of the ILS metaheuristic. The eILS combines a stack of local search operators, removal heuristics, a best insertion algorithm, and an intensification/diversification mechanism based on an \textit{elite set} of solutions. The performance of the proposed method was compared with that of a math-heuristic approach from the literature. The eILS achieved excellent results by improving the best-known solution for many benchmark instances. 

Several promising research perspectives was determined after the study of the TRSP. This study highlights the need to design exact approaches for TRSP. Mathematical models, either linear or non-linear, seem to be not suitable and often fail to solve small instances. This is mainly because of the nature of the objective function, where instead of minimizing travel costs, the duration is minimized. The most suitable approach for finding optimal solutions for the TRSP is the branch-and-price method \cite{desrochers1988}. Another promising direction is the integration of the synchronization constraints into the TRSP. This allows the problem to cover relevant cases where a given task requires the intervention of multiple technicians, either simultaneously or with precedence relations, to perform the maintenance operation. Finally, because the objective function aims at the minimization of the total duration, a promising research direction consists in the elaboration of methods that consider workload balancing.

{
\footnotesize
\section*{Acknowledgment}
This work was carried out within the framework of the ELSAT220 project. The ELSAT2020 project is co-financed by the Hauts-de-France Region and European Economic and Regional Development Fund (ERDF) of the EU.
}

\section*{CRediT authorship contribution statement}
\textbf{Ala-Eddine Yahiaoui:} Conceptualization, Methodology, Software, Validation, Writing – original draft, Writing – review \& editing. \textbf{Sohaib Afifi and Hamid Allaoui:} Review \& editing.

\bibliographystyle{plain}
\bibliography{biblio}

\appendix

\section{Detailed Results}\label{app:detailed}

{
 \scriptsize
\begin{longtable}[h]{lccccc}
   \caption{\sc Detailed results of pALNS and eILS}\\
   \toprule
   \multirow{2}{*}{Instance} & \multirow{2}{*}{Best} & pALNS & \multicolumn{3}{c}{eILS}\\
	\cmidrule(lr){3-3}\cmidrule(lr){4-6} & &
     $BEST$ & $BEST$ & $AVG$ & $CPU$\\
   \midrule
   \endfirsthead
  
   \multicolumn{6}{c}{{\tablename} \thetable{} -- continued from previous page} \\
  
   \toprule
   \multirow{2}{*}{Class}& \multirow{2}{*}{Best} & pALNS & \multicolumn{3}{c}{eILS}\\
   
	\cmidrule(lr){3-3}\cmidrule(lr){4-6} & &
      $BEST$& $BEST$ & $AVG$ & $CPU$\\
\midrule
  
\endhead
  
\multicolumn{6}{c}{continued on next page}
\endfoot
\bottomrule
\endlastfoot

$C101$&$10685.9$&$10717.52$&\textbf{$10685.9$}&$10743.4$&$245.649$\\
$C102$&$10228.8$&$10239.04$&\textbf{$10228.8$}&$10262.7$&$391.555$\\
$C103$&$10281.86$&\textbf{$10281.86$}&$10285.6$&$10308.5$&$338.768$\\
$C104$&$10107.44$&\textbf{$10107.44$}&$10120.6$&$10136.9$&$354.258$\\
$C105$&$10570.1$&$10584.06$&\textbf{$10570.1$}&$10592.6$&$271.371$\\
$C106$&$10321$&$10322.6$&\textbf{$10321$}&$10355.4$&$296.845$\\
$C107$&$10356.56$&\textbf{$10356.56$}&$10370.2$&$10419.1$&$322.745$\\
$C108$&$10251.22$&\textbf{$10251.22$}&$10260.1$&$10282.3$&$372.89$\\
$C109$&$10107.33$&\textbf{$10107.33$}&$10117.4$&$10132$&$375.688$\\
$C201$&$10188.4$&$10192.98$&\textbf{$10188.4$}&$10200.2$&$298.42$\\
$C202$&$9920.53$&$10001.52$&\textbf{$9920.53$}&$9923$&$401.808$\\
$C203$&$9962.36$&$10001.17$&\textbf{$9962.36$}&$9978.06$&$410.631$\\
$C204$&$9858.65$&$9890.56$&\textbf{$9858.65$}&$9869.22$&$469.463$\\
$C205$&$10142.6$&$10208.44$&\textbf{$10142.6$}&$10161.4$&$365.68$\\
$C206$&$9918.08$&$9983.14$&\textbf{$9918.08$}&$9920.02$&$508.429$\\
$C207$&$9839.19$&$9849.69$&\textbf{$9839.19$}&$9847.17$&$411.334$\\
$C208$&$9898.81$&$9981.06$&\textbf{$9898.81$}&$9918.9$&$453.244$\\
$R101$&$3134.86$&\textbf{$3134.86$}&$3138.52$&$3162.75$&$270.218$\\
$R102$&$3034.3$&$3039.23$&\textbf{$3034.3$}&$3068.63$&$339.597$\\
$R103$&$2421.78$&\textbf{$2421.78$}&$2424.91$&$2454.15$&$417.04$\\
$R104$&$2277.65$&$2285.77$&\textbf{$2277.65$}&$2301$&$308.667$\\
$R105$&$2961.04$&$2975.42$&\textbf{$2961.04$}&$2991.34$&$346.455$\\
$R106$&$2626.38$&\textbf{$2626.38$}&$2636.44$&$2661.16$&$372.324$\\
$R107$&$2131.73$&$2134.36$&\textbf{$2131.73$}&$2137.79$&$481.441$\\
$R108$&$2116.05$&\textbf{$2116.05$}&$2116.24$&$2125.77$&$286.345$\\
$R109$&$2512.89$&\textbf{$2512.89$}&$2539.55$&$2563.16$&$342.538$\\
$R110$&$2359.42$&\textbf{$2359.42$}&$2384.74$&$2412.67$&$378.265$\\
$R111$&$2550.5$&\textbf{$2550.5$}&$2567.22$&$2595.9$&$278.235$\\
$R112$&$2145.76$&\textbf{$2145.76$}&$2163.34$&$2178.27$&$305.27$\\
$R201$&$2635.07$&$2639.7$&\textbf{$2635.07$}&$2654.05$&$352.272$\\
$R202$&$2382.87$&\textbf{$2382.87$}&$2390.13$&$2407.44$&$412.038$\\
$R203$&$2319.49$&$2334.58$&\textbf{$2319.49$}&$2345.65$&$350.529$\\
$R204$&$1931.19$&\textbf{$1931.19$}&$1939.22$&$1948.32$&$405.609$\\
$R205$&$2230.52$&$2254.82$&\textbf{$2230.52$}&$2248.26$&$354.095$\\
$R206$&$2061.3$&$2082.8$&\textbf{$2061.3$}&$2077.93$&$352.69$\\
$R207$&$1963.57$&$1981.76$&\textbf{$1963.57$}&$1968.31$&$401.502$\\
$R208$&$1829.73$&$1879.49$&\textbf{$1829.73$}&$1840.72$&$487.621$\\
$R209$&$2066.32$&$2130.5$&\textbf{$2066.32$}&$2098.59$&$327.945$\\
$R210$&$2062.32$&$2111.92$&\textbf{$2062.32$}&$2077.98$&$375.614$\\
$R211$&$1883.63$&$1975.42$&\textbf{$1883.63$}&$1907.86$&$382.419$\\
$RC101$&$2856.49$&\textbf{$2856.49$}&$2871.13$&$2903.32$&$296.846$\\
$RC102$&$2843.18$&\textbf{$2843.18$}&$2851.81$&$2883.94$&$307.664$\\
$RC103$&$2474.75$&$2495.57$&\textbf{$2474.75$}&$2508.13$&$323.021$\\
$RC104$&$2162.51$&\textbf{$2162.51$}&$2173.1$&$2193.21$&$362.273$\\
$RC105$&$2711.49$&\textbf{$2711.49$}&$2720.56$&$2731.9$&$350.05$\\
$RC106$&$2761.86$&\textbf{$2761.86$}&$2799.53$&$2816.57$&$279.961$\\
$RC107$&$2570.4$&\textbf{$2570.4$}&$2608.63$&$2635$&$337.183$\\
$RC108$&$2354.42$&\textbf{$2354.42$}&$2364.49$&$2397.79$&$362.306$\\
$RC201$&$2682.56$&$2686.01$&\textbf{$2682.56$}&$2722.5$&$352.518$\\
$RC202$&$2457.65$&$2487.73$&\textbf{$2457.65$}&$2498.4$&$390.661$\\
$RC203$&$2283.23$&$2310.01$&\textbf{$2283.23$}&$2320.93$&$362.624$\\
$RC204$&$2016.53$&$2064.16$&\textbf{$2016.53$}&$2030.11$&$399.336$\\
$RC205$&$2544.14$&$2588.17$&\textbf{$2544.14$}&$2582.74$&$355.996$\\
$RC206$&$2321.21$&$2359.7$&\textbf{$2321.21$}&$2341.35$&$372.752$\\
$RC207$&$2176.09$&$2233.68$&\textbf{$2176.09$}&$2191.61$&$367.595$\\
$RC208$&$1895.13$&$1914.54$&\textbf{$1895.13$}&$1903.46$&$351.896$\\

\hline
$Average$ & $4739.09$ & $4758.63$ & $4744.39$ & $4766.78$ & $360.54$\\

\label{tab:comparisonPillac}
\end{longtable}
}

\end{document}